\documentclass[letterpaper]{article} 
\usepackage[submission]{aaai24}  
\usepackage{times}  
\usepackage{helvet}  
\usepackage{courier}  
\usepackage[hyphens]{url}  
\usepackage{graphicx} 
\usepackage{natbib}  
\usepackage{caption} 
\usepackage{algorithm}
\usepackage{algorithmic}
\usepackage{booktabs}

\usepackage{subfigure,color}
\usepackage{multirow,amsmath}
\usepackage{array}
\usepackage{amssymb}
\usepackage{multirow,amsmath}
\usepackage{tabulary}
\usepackage{pgfplots}

\usepackage{newfloat}
\usepackage{listings}
\usepackage{dsfont}

\DeclareCaptionStyle{ruled}{labelfont=normalfont,labelsep=colon,strut=off} 
\floatstyle{ruled}
\newfloat{listing}{tb}{lst}{}
\floatname{listing}{Listing}
\title{LRANet: Towards Accurate and Efficient Scene Text Detection with Low-Rank Approximation Network}
\author {
    Yuchen Su\textsuperscript{\rm 1,2},
    Zhineng Chen\textsuperscript{\rm 1}\footnote{Corresponding author.},
    Zhiwen Shao\textsuperscript{\rm 3},
    Yuning Du\textsuperscript{\rm 2},
    Zhilong Ji\textsuperscript{\rm 4}
    Jinfeng Bai\textsuperscript{\rm 4}
    Yong Zhou\textsuperscript{\rm 3}
    Yu-Gang Jiang\textsuperscript{\rm 1}
}
\affiliations {
    \textsuperscript{\rm 1} Shanghai Collaborative Innovation Center of Intelligent Visual Computing, School of Computer Science, Fudan University \\ 
    \textsuperscript{\rm 2} Baidu Inc. \\
    \textsuperscript{\rm 3} China University of Mining and Technology\\
    \textsuperscript{\rm 4} Tomorrow Advancing Life\\
    ycsu23@m.fudan.edu.cn,
    \{zhinchen, ygj\}@fudan.edu.cn, duyuning@biadu.com, jizhilong@tal.com, jfbai.bit@gmail.com, \{zhiwen\_shao, yzhou\}@cumt.edu.cn
}

\usepackage{bibentry}

\begin{document}

\maketitle

\begin{abstract}
Recently, regression-based methods, which predict parameterized text shapes for text localization, have gained popularity in scene text detection. However, the existing parameterized text shape methods still have limitations in modeling arbitrary-shaped texts due to ignoring the utilization of text-specific shape information. Moreover, the time consumption of the entire pipeline has been largely overlooked, leading to a suboptimal overall inference speed. 
To address these issues, we first propose a novel parameterized text shape method based on low-rank approximation. 
Unlike other shape representation methods that employ data-irrelevant parameterization, our approach utilizes singular value decomposition and reconstructs the text shape using a few eigenvectors learned from labeled text contours. By exploring the shape correlation among different text contours, our method achieves consistency, compactness, simplicity, and robustness in shape representation. Next, we propose a dual assignment scheme for speed acceleration. It adopts a sparse assignment branch to accelerate the inference speed, and meanwhile, provides ample supervised signals for training through a dense assignment branch. Building upon these designs, we implement an accurate and efficient arbitrary-shaped text detector named LRANet. Extensive experiments are conducted on several challenging benchmarks, demonstrating the superior accuracy and efficiency of LRANet compared to state-of-the-art methods. Code is available at:  \url{https://github.com/ychensu/LRANet.git}
\end{abstract}

\section{Introduction}

Scene text detection has garnered significant attention in computer vision due to its fundamental role as the initial step for end-to-end text recognition \cite{sheng2019nrtr,du2022svtr,zheng2023tps++}. 
Driven by the rapid development of deep learning, scene text detection methods have made significant progress \cite{song2022unsupervised,wang2022petr,fang2022abinet++,song2022unsupervised,chen2022date,song20233d,qin2023towards,zheng2023cdistnet,du2023context,wang2023real,shu2023perceiving}.

\begin{figure}[t]
    \centering
    \includegraphics[width=\linewidth]{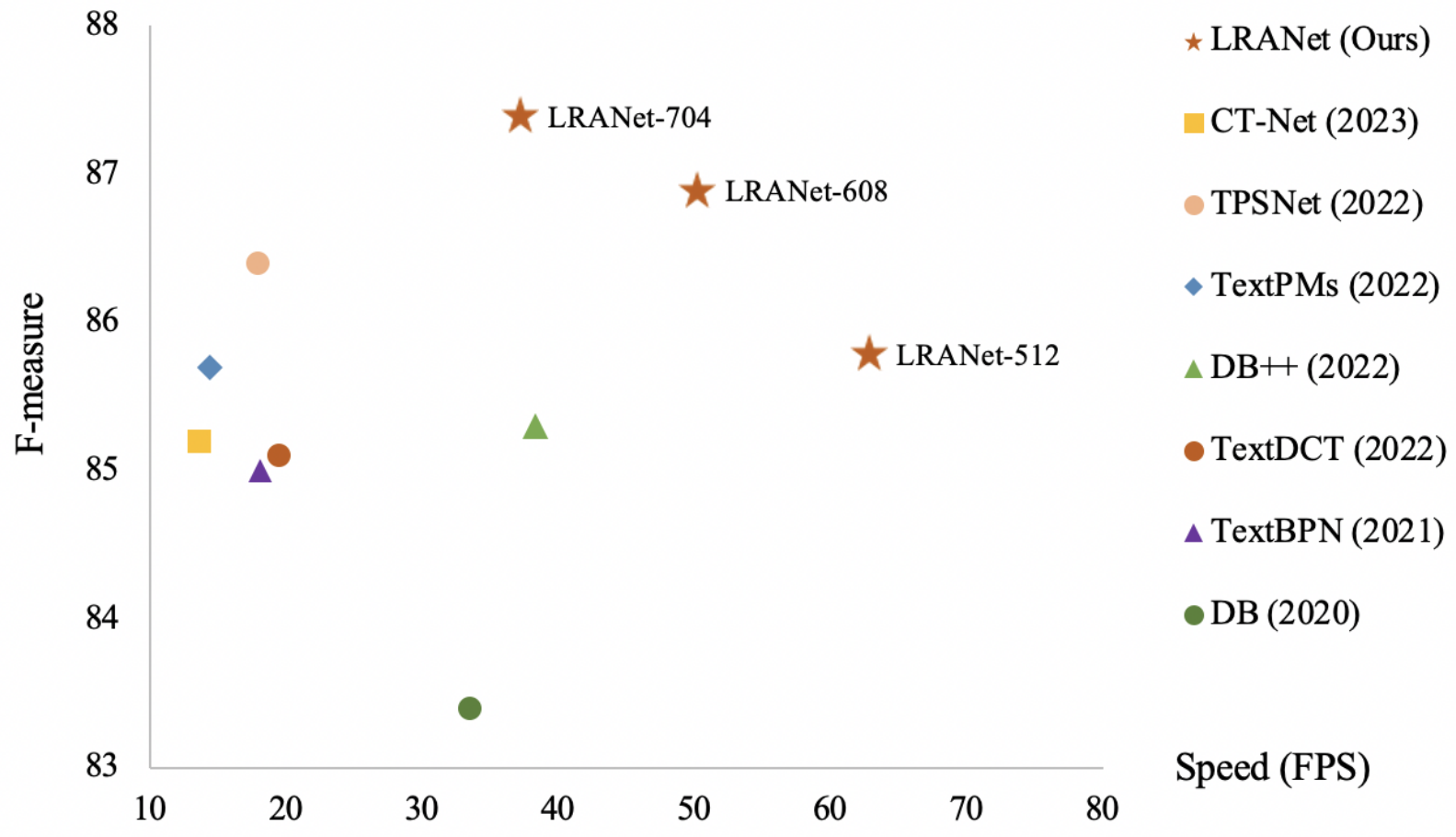}
    \caption{The comparisons of several popular scene text detectors on the CTW1500 dataset. Our method achieves the best trade-off between accuracy and efficiency.}
    \label{fig-efficient-accurate}
\end{figure}

Existing scene text detection can be roughly divided into segmentation-based methods \cite{wang2019arbitrary,zhou2020crnet,zhu2021textmountain,liao2022real} and regression-based methods \cite{sheng2019single,hou2020ham,wang2020contournet,zhang2022kernel,su2022textdct}. 
To explore an arbitrary-shaped scene text detector, segmentation-based methods model text instances with pixel-level classification masks that allow naturally fitting of arbitrary shapes, but they mainly focus on local textual cues rather than the integrated geometric layout of texts, leading to a lack of global perception. Complex post-processing is often required to merge the pixel-level results to text regions.

Recently, regression-based methods, which predict parameterized text shapes for text localization, have sparked a new wave of research in arbitrary-shaped text detection due to their enhanced ability to consider the integrated geometric layout of texts. However, there are still two main problems that remain to be explored.

\begin{figure}[t]
    \centering
    \subfigure [TextRay \cite{wang2020textray}]
    {
        \includegraphics[width=0.22\textwidth,height=2.0cm]{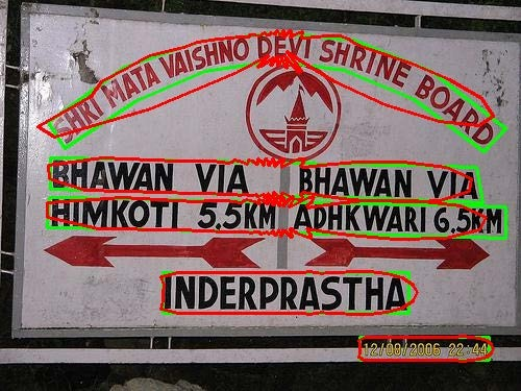}
        \label{fig-TextRay}
    }
    \subfigure [TextDCT \cite{su2022textdct}]
    {
        \includegraphics[width=0.22\textwidth,height=2.0cm]{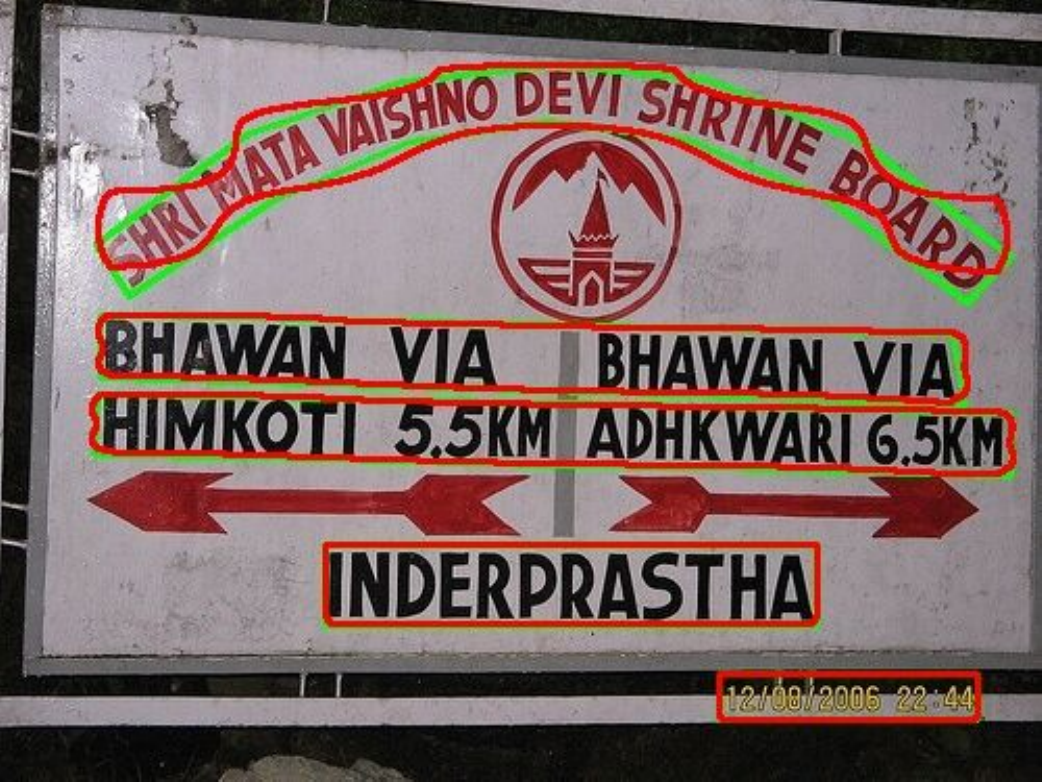}
        \label{fig-TextDCT}
    }
    \subfigure [FCENet \cite{zhu2021fourier}]
    {
        \includegraphics[width=0.22\textwidth,height=2.0cm]{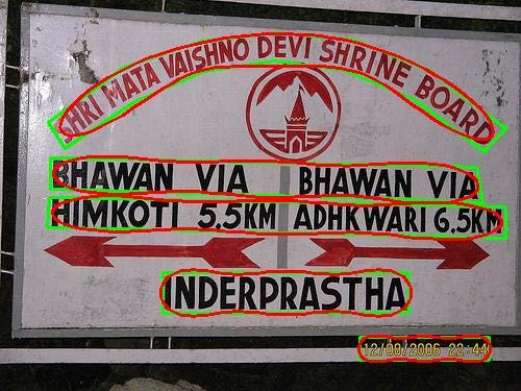}
        \label{fig-FCENet}
    }
    \subfigure [\textbf{Ours}]
    {
        \includegraphics[width=0.22\textwidth,height=2.0cm]{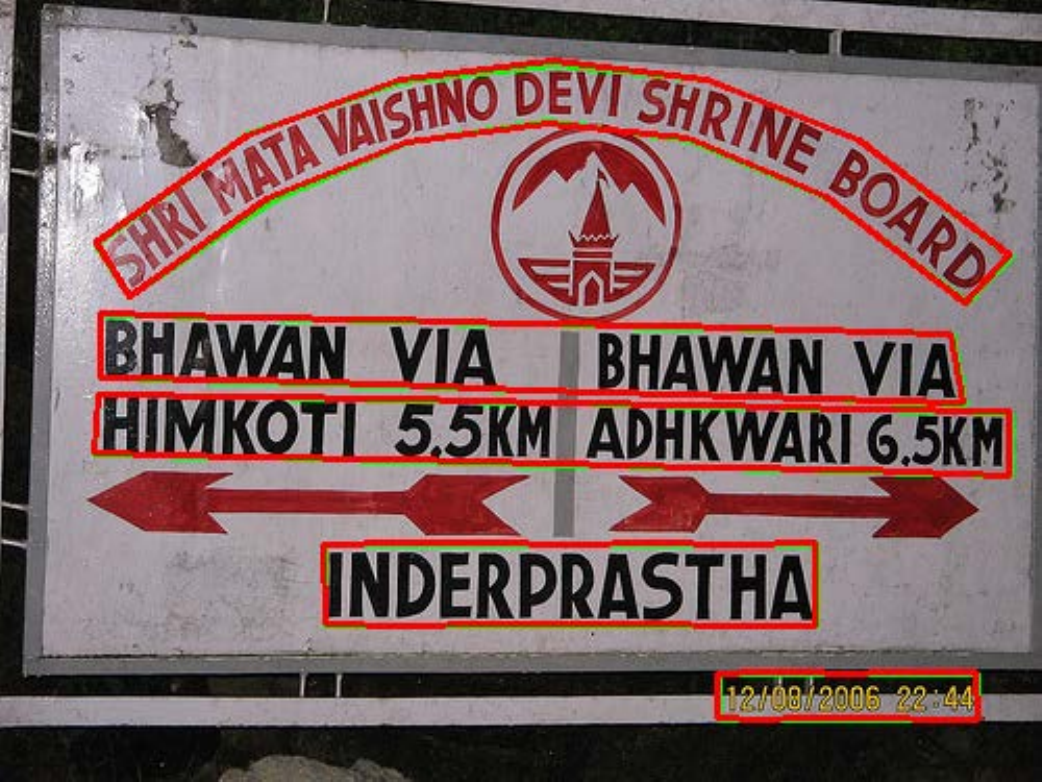}
        \label{fig-lranet}
    }
    \caption{Comparison of different parameterized text shape methods. Ground-truth contours are depicted in green, and the fitted curves are shown in red. The number of fitting parameters are $44$, $32$, $22$ and $\mathbf{14}$ from (a) to (d). Ours gets a superior contour representation using fewer parameters.}
    \label{fig-compare}
\end{figure}

One problem is that the existing parameterized text shape methods are still challenging in modeling arbitrary-shaped texts. For horizontal and multi-oriented text, regressing quadrilaterals is sufficient to represent the text shape. However, sophisticated representation is required for representing arbitrary-shaped text. Some methods adopt contour points or parametric curves to fit the text shape, but they either lack imposing geometric constraints or fail to consider the distinct characteristics of text shapes. As a result, these methods may not faithfully represent the text boundaries, as shown in Fig.~\ref{fig-TextRay}, ~(b) and ~(c). Scene text exhibits a wide range of shape diversity and aspect ratios. However, current parameterized text shape methods solely model text shapes individually using data-irrelevant decomposition, ignoring the structural relationships among different text shapes and the utilization of text-specific shape information. This makes it challenging to consistently and robustly represent various text shapes using only a few parameters.

The other problem is that regression-based methods often overlook the overall speed of the entire pipeline. Specifically, current regression-based methods can be divided into dense assignment approaches
\cite{liu2020abcnet,su2022textdct,wang2022tpsnet} and one-to-one assignment approaches \cite{raisi2022arbitrary, ye2023dptext, shao2023ct} according to the allocation of positive samples. However, dense assignment approaches require 
non-maximum suppression (NMS) to filter a large number of redundant predictions, and this process is time-consuming, especially for arbitrary-shaped redundant predictions. Although one-to-one assignment approaches adopt the set prediction mechanism from DETR \cite{carion2020end} to address this issue, they lack sufficient supervised signals and explicit position prior information. As a result, it usually stacks multiple decoders for iterative text contour optimization, leading to a complex pipeline.

To tackle the above-mentioned problems, firstly,
inspired by the recent instance segmentation work \cite{park2022eigencontours}, we propose a low-rank approximation (LRA) method to better represent arbitrary-shaped texts. Unlike previous parameterized text shape methods that solely consider the individual text shape information, our LRA learns to represent text contours by exploring the shape correlation among different text instances. In detail, we first construct a text contour matrix, which contains all text contours in the training set. Then, we apply singular value decomposition (SVD) to extract the first $M$ left singular vectors of the contour matrix, referred to as eigenanchors. Based on the best $M$-approximation property of SVD, each text contour can be approximated as a linear combination of the eigenanchors, as illustrated in Fig.~\ref{fig-overview}. LRA endows the generated shape representation with the following characteristics: compactness (fewer fitting parameters), consistency (fitting various text shapes tightly), simplicity (fitting with only linear operations). Moreover, LRA is robust, as the contour is mainly determined by the top-ranked eigenanchors while the subsequent ones only affect the edge detail.

Next, we propose a dual assignment scheme for positive samples, aiming to accurately learn and efficiently infer the coefficients of eigenanchors. Specifically, we introduce a sparse assignment branch to reduce redundant predictions and thus decrease the processing time of NMS during inference, while employing a dense assignment branch to provide ample supervised signals.

Building upon these designs, we develop an accurate and efficient scene text detector named LRANet, whose performance advantages are shown in Fig.~\ref{fig-efficient-accurate}. The main contributions of this paper are summarized as follows:

\begin{figure}[t]
\centering
\includegraphics[width=\linewidth]{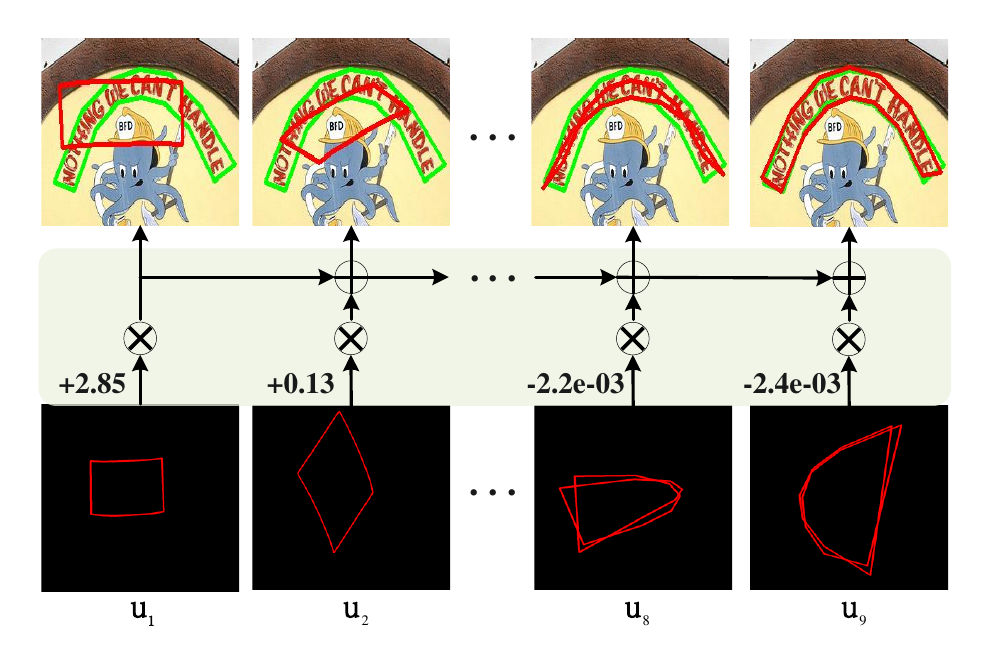}
\caption{Illustration of the low-rank approximation representation. The ground-truth contour is depicted in green, with $u_1$,  $u_2$, ...,  $u_8$ and $u_9$ as eigenanchors. The text contour is approximated by a linear combination of the eigenanchors, and fitted curves using different numbers of eigenanchors are shown from left to right.
}
\label{fig-overview}
\end{figure}

    $\bullet$ We propose a novel parameterized text shape method named LRA. It represents text shapes accurately by leveraging a linear combination of eigenvectors learned from arbitrary-shaped text contours.

    $\bullet$ We introduce a dual assignment scheme by complementarily combining sparse and dense assignments. It improves the model's inference efficiency while ensuring its accuracy.
    
    $\bullet$ We propose a light-weighted regression-based text detector named LRANet, which can accurately and efficiently detect arbitrary-shaped texts. Extensive experiments show that LRANet achieves state-of-the-art performance.

\section{Related Work}
\subsection {Segmentation-Based Methods}
Segmentation-based methods regard the text detection as a segmentation problem, firstly modeling text instances with pixel-level classification masks, and then formed them into text boundaries through specific heuristic operations. PSENet \cite{wang2019Shape} predicts text instances with different scale kernels and applies a progressive scale expansion strategy to gradually expand the predefined text kernels. DB \cite{liao2020Real} and DB++ \cite{liao2022real} introduce a differentiable binarization module that assigns a higher threshold to text boundaries, enabling the distinction between adjacent text instances. 
TextPMs \cite{zhang2022arbitrary} proposes an iterative model to predict a set of probability maps, and then utilizes region growth algorithms to group these probability maps into text instances.

Unfortunately, these segmentation-based methods model text instances from a local perspective, lacking global perception of texts, resulting in sensitivity to text-like background noise and computationally intensive post-processing.

\begin{figure*}
\centering
\includegraphics[width=0.8\textwidth]{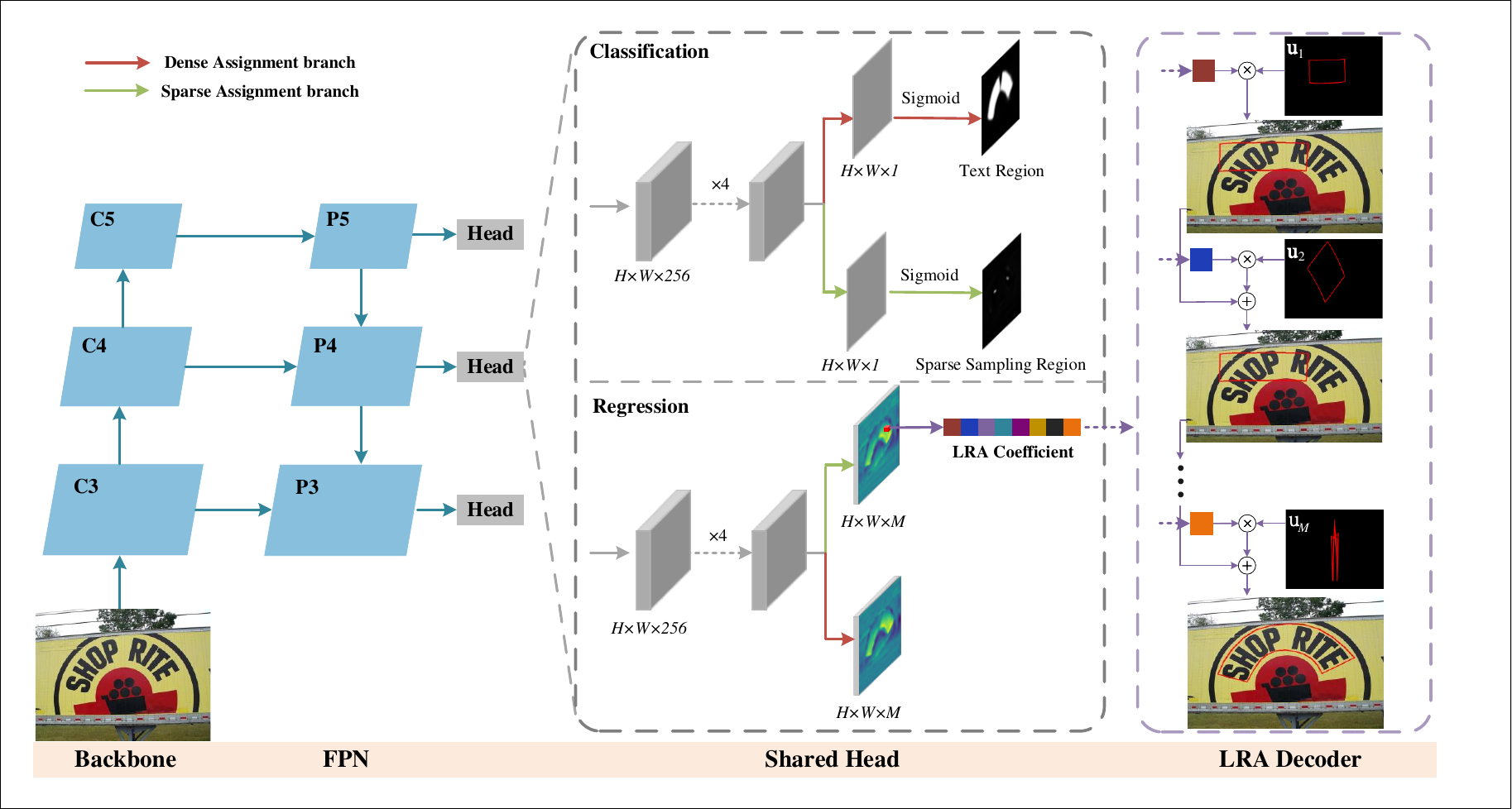}
\caption{The overview of our LRANet, which is mainly composed of three modules: (a) the backbone and FPN for feature extraction, (b) shared head for joint optimization, and (c) LRA decoder to reconstruct the text shape. The classification branch and the regression branch are used for predicting the positive samples and LRA coefficients, respectively.}
\label{fig-framwork}
\end{figure*}

\subsection {Regression-Based Methods}
Regression-based methods are mainly inspired by the advancement in object detection, where text shapes are represented as vectors through parameterization methods for regression. These methods primarily focus on irregular texts due to their complex representation. Earlier approaches directly regress contour points as text boundary, but they fail to utilize prior information about the continuity of text boundary. Therefore, later approaches employ parameterized curves or parameterized masks to represent the text boundary. For example, TextRay \cite{wang2020textray} adopts Chebyshev polynomials under the polar system to approximate the text boundary. 
ABCNet \cite{liu2020abcnet} adopts Bernstein polynomial to convert the long sides of texts into Bezier curves.
FCENet \cite{zhu2021fourier} converts text contours to Fourier signature vectors by Fourier contour embedding. TextDCT \cite{su2022textdct} transforms text instance masks to the Frequency domain by discrete cosine transform (DCT), and then extracts the low-frequency components to represent the text instance masks. TPSNet \cite{wang2022tpsnet} utilizes thin plate splines (TPS) to parameterize text contours into TPS fiducial points.

However, these parameterized methods have not sufficiently taken into account text-specific shape information, leading to limitations in modeling arbitrary-shaped texts, For example, chebyshev polynomials and DCT representation struggle to tightly fit the text shape with a compact vector. Fourier representation may lose some corner pixels, as shown in Fig.~\ref{fig-compare}. Although TPS and Bezier can fit long curved texts well through fiducial points, slight perturbations in these points would significantly affect the fitted curve. Meanwhile, using a sparse number of fiducial points may not be sufficient to accurately represent an arbitrary-shaped text. Different from these efforts, we propose LRA that represents the fitted curve from an eigenvector-reconstruction perspective, allowing for effective exploitation of text-specific shape information. 

\section{Methodology}

\subsection {Low-Rank Approximation Representation}

Low-rank approximation (LRA) is a widely-applied data dimensional reduction method.  
In this paper, we use LRA to compactly represent text contours. Unlike previous parameterized text shape methods that use curve fitting \cite{liu2020abcnet,zhu2021fourier,wang2022tpsnet} or mask compression \cite{su2022textdct}, LRA is a data-driven approach that represents text boundaries in a low-dimensional space by exploiting the distribution of labeled text contours.

Scene text shapes are typically well-structured, mainly characterized by large aspect ratios and right-angle corners. As a result, there is significant correlation among these text shapes. By exploiting this correlation based on labeled data, we design a novel parameterized text shape method. Specifically, the ground-truth text boundary typically consists of multiple vertices, we first flatten them into a column vector $\mathbf{p}=\left[\mathbf{x}_1, \mathbf{y}_1, \cdots, \mathbf{x}_N,\mathbf{y}_N\right]^{\top}\in{\mathbb{R}^{{2N} \times 1}}$, where $N$ is the number of vertices. Then, we construct a text contour matrix $\mathbf{A}=\left[\mathbf{p}_1, \mathbf{p}_2, \cdots, \mathbf{p}_L\right]\in{\mathbb{R}^{{2N} \times L}}$ from a corpus containing $L$ labeled text instances. Thus, we apply singular value decomposition (SVD) to extract the structural relationship among the text contours in $\mathbf{A}$ to
\begin{equation}
\label{eq-A=u...}
\mathbf{A}=\mathbf{U} \boldsymbol{\Sigma} \mathbf{V}^{\top}.
\end{equation}
Here, $\mathbf{U}=\left[\mathbf{u}_1, \mathbf{u}_2, \cdots, \mathbf{u}_{2N }\right]\in{\mathbb{R}^{{2N} \times 2N}}$ and $\mathbf{V}=\left[\mathbf{v}_1, \mathbf{v}_2, \cdots, \mathbf{v}_{L}\right]\in{\mathbb{R}^{{L} \times L}}$ are orthogonal matrices, and    $\boldsymbol{\Sigma}\in{\mathbb{R}^{{2N} \times L}}$ is a diagonal matrix composed of singular values $\sigma_1 \geq \sigma_2 \geq \cdots \geq \sigma_r>0$, where $r$ denotes the rank of $\mathbf{A}$. Therefore, we can decompose $\mathbf{A}$ as
\begin{equation}
\label{eq-A=sigmauv}
\mathbf{A}=\sigma_1 \mathbf{u}_1 \mathbf{v}_1^{\top}+\cdots+\sigma_r \mathbf{u}_r \mathbf{v}_r^{\top},
\end{equation}
where $\mathbf{u}_i$ and $\mathbf{v}_i$ are left singular and right singular vectors corresponding to $\sigma_i$, respectively.

After SVD, each singular vector represents an orthogonal direction in the latent space. 
By retaining only the first $M$ singular values and their corresponding singular vectors, and truncating the low-rank components obtained from the decomposition in Eq.~\eqref{eq-A=sigmauv}, matrix $\mathbf{A}$ can be approximated by
\begin{equation}
\begin{aligned}
 \mathbf{A}_M & = \sigma_1 \mathbf{u}_1 \mathbf{v}_1^{\top}+\cdots+\sigma_M \mathbf{u}_M \mathbf{v}_M^{\top}=\mathbf{U}_M \boldsymbol{\Sigma}_M \mathbf{V}_M^{\top} \\ &
=\left[\tilde{\mathbf{p}}_1, \cdots, \tilde{\mathbf{p}}_L\right] \approx \mathbf{A},
\end{aligned}
\end{equation}
where $\tilde{\mathbf{p}}_i$ denotes the approximation of $\mathbf{p}_i$, and $\mathbf{A}_M$ is the best $M$-rank approximation of $\mathbf{A}$ as it minimizes the Frobenius norm $|\mathbf{A}-\mathbf{A}_M\|_F$  \cite{blum2015foundations}.

Then, we define $\boldsymbol{\Sigma}_M \mathbf{V}_M^{\top}$ as matrix $\mathbf{C}_M=\left[\mathbf{c}_1, \mathbf{c}_2, \cdots, \mathbf{c}_{L}\right]\in{\mathbb{R}^{{M} \times L}}$, allowing the matrix $\mathbf{A}_M$ to be represented as
\begin{equation}
\label{eq-Am=UmC}
 \mathbf{A}_M =  \left[{\mathbf{U}_M \mathbf{c}}_1, \cdots, {\mathbf{U}_M \mathbf{c}}_L\right]=\left[\tilde{\mathbf{p}}_1, \cdots, \tilde{\mathbf{p}}_L\right].
\end{equation}
In Eq.~\eqref{eq-Am=UmC}, each approximated text contour $\tilde{\mathbf{p}_i}$ can be expressed as a linear combination of the first $M$ left singular vectors $\mathbf{u}_1, \mathbf{u}_2, \cdots, \mathbf{u}_{M}$. In other words,
\begin{equation}
\label{eq-p=uc}
\tilde{\mathbf{p}}_i=\mathbf{U}_M \mathbf{c}_i=\left[\mathbf{u}_1, \cdots, \mathbf{u}_M\right] \mathbf{c}_i.
\end{equation}

We call these $\mathbf{u}_1, \cdots, \mathbf{u}_M$
as $eigenanchors$, as they are eigenvectors of matrix $\mathbf{A} \mathbf{A}^{\top}$ and can be viewed as pre-defined arbitrary-shaped anchors, as shown in Fig.~\ref{fig-overview}. Worth noting, eigenanchors are similar to the principal components in principal components analysis (PCA). However, PCA eliminates the mean value during matrix approximating, which may lead to information loss in our case.

We refer to the space spanned by the eigenanchors as $eigenanchor$ $space$.    
For any $2N$-dimensional text contour $\mathbf{p}$, we can project it onto the eigenanchor space to obtain an $M$-dimensional representation by
\begin{equation}
{\mathbf{c}}=\mathbf{U}_M^{\top} \mathbf{p},
\end{equation}
where $\mathbf{c}$ is a coefficient vector. Also, the approximation of $\mathbf{p}$ can be reconstructed by Eq.~\eqref{eq-p=uc}.
Note that the number of contour vertices in $\mathbf{p}$ may not equal to $N$. In that case, we resample $N$ vertices from $\mathbf{p}$ using cubic spline interpolation.

\subsection {Dual Assignment Scheme}
We propose a dual assignment scheme for positive samples. It adopts a dense assignment branch to provide ample supervised signals for training, and meanwhile, accelerates the inference speed through a sparse assignment branch, enjoying the merits of both branches. 

Specifically, following previous regression-based text detection network \cite{zhu2021fourier,wang2022tpsnet},
our dense assignment scheme treats the text region as the positive sample region. In our sparse assignment scheme, 
we construct a prediction-aware matrix $\mathbf{S}$ for selecting $K$ positive samples for each text instance, 
The matrix element is defined as 
\begin{equation}
\label{eq-Selection-metric}
{s}_{i j}=\left\{\begin{array}{cc}
 \mathrm{FL}^{\prime}({b}_{i}) + \lambda \sum\limits_{n=0}^{N-1}\left\|{\tilde{\mathbf{p}}}_{\raisebox{-0.2ex}{$\scriptstyle i$}}^{(n)}-\mathbf{p}_{\raisebox{-0.2ex}{$\scriptstyle j$}}^{(n)}\right\|,    & i \in TR \\[3ex]
 \infty, & i \notin TR
\end{array} . \right .
\end{equation}
Here, ${s}_{i, j}$ denotes the matching cost between the $i$-th point and the $j$-th ground-truth text instance, ${b}_{i}$ is the predicted classification score of the $i$-th point. $\mathrm{FL}^{\prime}$ is defined as the difference between the positive and negative terms: $\mathrm{FL}^{\prime} = -\alpha(1-x)^\gamma \log (x)+(1-\alpha) x^\gamma \log (1-x)$, which is derived from the focal loss \cite{lin2017focal}. We set $\alpha$ to $0.25$ and $\gamma$ to $2.0$. The second term is the L1 distance between the $i$-th predicted contour and $j$-th 
ground-truth contour, and $\lambda$ controls the importance degree of classification and regression. The third item aims to limit the sparse sampling to only the text region for better joint optimization. Note that both the cost of classification and regression are considered in Eq.~\eqref{eq-Selection-metric}.

Afterwards, we regard the sparse positive sampling as a bipartite matching problem and use the Hungarian algorithm to solve the matrix $\mathbf{S}$ in ascending order, to find the optimal matching point for each text instance. To explore the optimal number of positive sample allocations for each text instance, we replicate it $K-1$ times when constructing the matrix $\mathbf{S}$, and thus assign $K$ positive samples to each instance.

In summary, the dense assignment branch assists in training the sparse assignment branch through providing ample supervised signals. During inference, we utilize sparse positive samples predicted by the sparse assignment scheme to accelerate post-processing.

\subsection {LRANet}

\subsubsection {Network Architecture}

As shown in Fig.~\ref{fig-framwork}, we adopt a compact one-stage fully-convolutional architecture for effective and efficient detection. The architecture mainly comprises three parts: a feature extraction module, a detection head, and an LRA decoder for inference. Specifically, in the feature extraction module, we utilize ResNet50 \cite{he2016deep} with DCN \cite{zhu2019deformable} as our backbone network, and adopt Feature Pyramid Network (FPN) \cite{lin2017feature} to extract multi-scale feature maps. The detection head has two branches, responsible for classification and regression respectively, attached to the output of feature map P3, P4 and P5 of FPN. In the classification branch, we adopt four $3 \times 3$ convolutions to extract classification features, and then employ individual $3 \times 3$ convolutions to predict the text region (TR) and sparse sampling region (SSR) respectively, where the SSR is the positive sample region in our sparse assignment scheme. Similarly, in the regression branch, we adopt four $3 \times 3$ convolutions to extract regression features, and then employ two parallel convolutional layers, each consisting of one $3 \times 3$  convolution, to predict the coefficients of LRA corresponding to TR and SSR respectively.

In the LRA decoder, we first perform pixel-wise multiplication of the predictions from TR and SSR to obtain the final positive samples. Then, we decode the LRA coefficients corresponding to these samples to reconstruct the text shape with the pre-defined eigenanchors, and the decoder process is defined by Eq.~\eqref{eq-p=uc}.

\subsubsection {Overall Loss}

In LRANet, the overall loss is formulated as
\begin{equation}
\label{loss1}
\mathcal{L}=\mathcal{L}_{cls}+ \mathcal{L}_{reg}, 
\end{equation}
where $L_{cls}$ and $L_{reg}$ are the losses for the classification
branch and the regression branch, respectively.

The classification loss consists of the text region loss  $\mathcal{L}_{tr}$ and the sparse sampling region loss $\mathcal{L}_{ssr}$:
\begin{equation}
\label{loss2}
\mathcal{L}_{cls}=\mathcal{L}_{tr}+ \mathcal{L}_{ssr},  
\end{equation}
where $\mathcal{L}_{tr}$ and $\mathcal{L}_{ssr}$ are the cross entropy loss and focal loss, respectively.

The regression loss is defined as
\begin{equation}
\mathcal{L}_{reg}=\mathds{1}^{da} \sum_{i}^{N_{da}} l_1\left(\tilde{\mathbf{p}}_{i}, \mathbf{p}_{i}\right),
\end{equation}
where $da$ is the positive sample region in our dual assignment scheme, $\mathds{1}$ is a spatial indicator, outputting $1$ when the point is within $da$ and $0$ otherwise, and $l_1$ is smooth-L1 loss.

\begin{figure}[t]
\centering
\includegraphics[width=\linewidth]{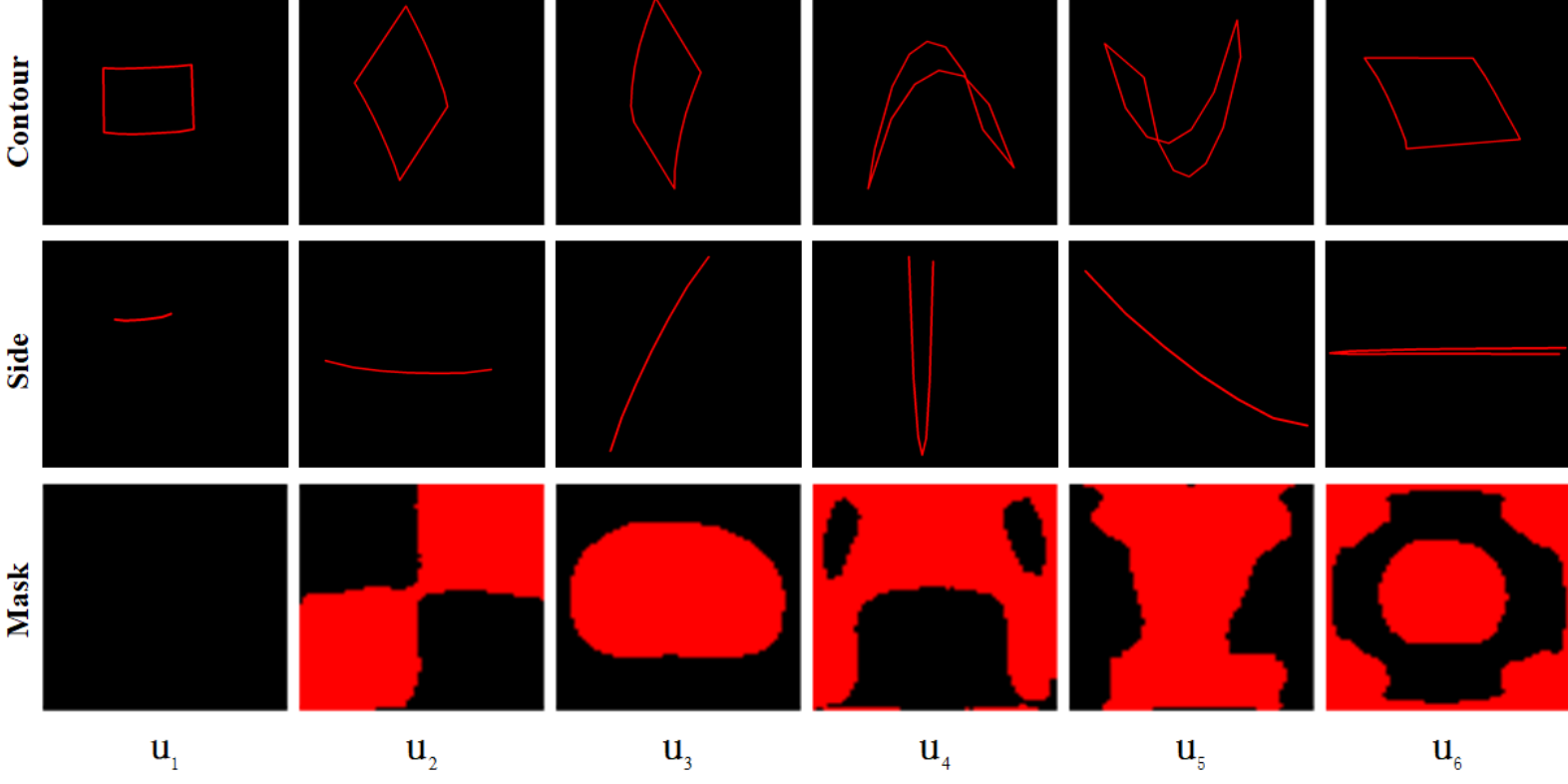}
\caption{Visualization of the first six eigenanchors with different data representations. These eigenanchors are obtained from the CTW1500 training dataset. 
}
\label{fig-mask-line-polygon}
\end{figure}

\begin{figure}[t]
\centering
\includegraphics[width=0.8\linewidth]{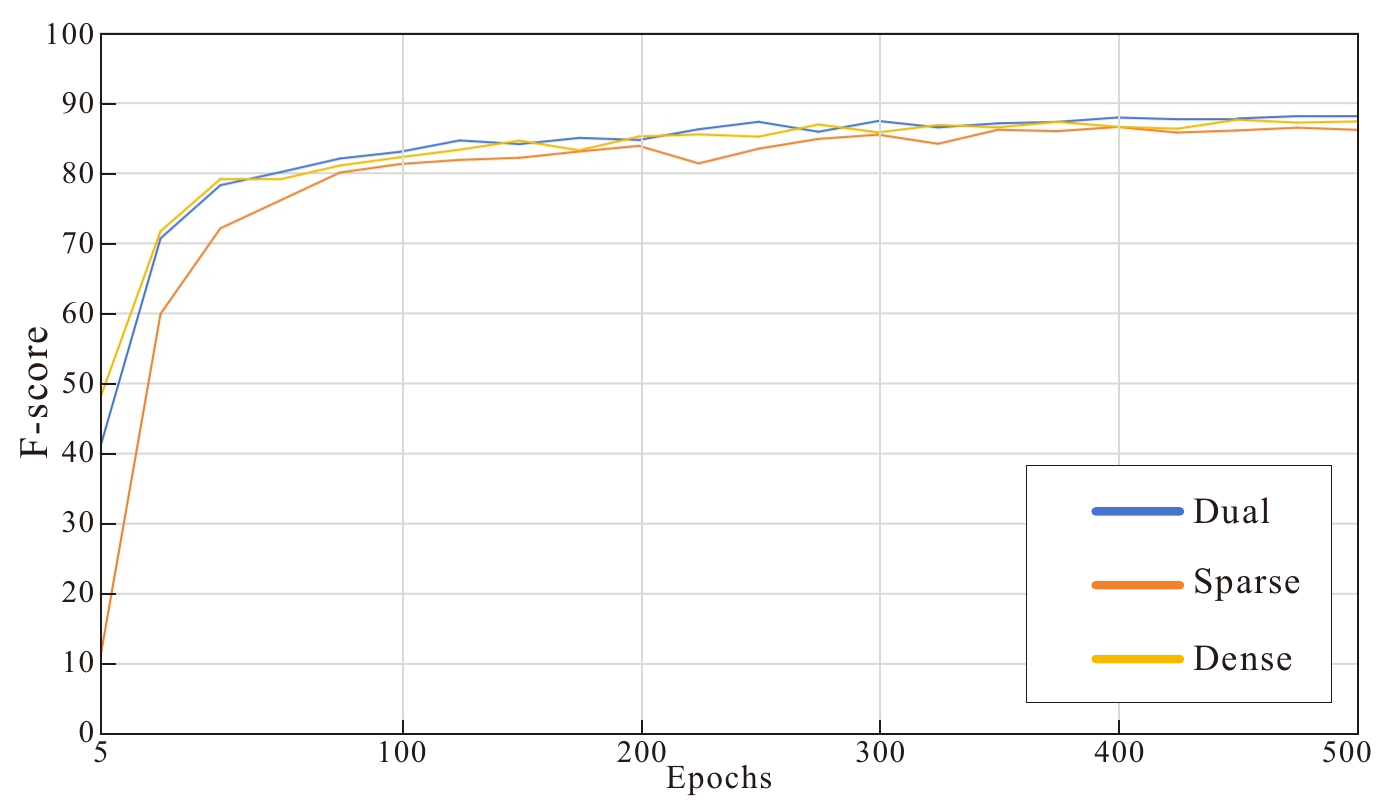}
\caption{Convergence curves of different assignment schemes on Total-Text.
}
\label{fig-convergence}
\end{figure}
\begin{table}[h]
\centering
\begin{tabular}{@{\extracolsep{10pt}}ccccc}
\hline
Data-Driven   &  R &  P     &  F    &  IoU     \\ \hline
Mask  & $84.5$   & $87.0$ & $85.7$ & $90.8$\\ 
Side   & $84.6$   & $88.6$          & $86.6$ & $97.9$ \\  
Contour  & $\mathbf{84.9}$   & $\mathbf{89.1}$ & $\mathbf{86.9}$ & $\mathbf{98.0}$ \\ 
\hline
\end{tabular}
\caption{Experimental results on CTW1500 for different data representation. IoU refers to the intersection over union between reconstructed text region and ground-truth text region.}
\label{tabel-mask-side-contour}
\end{table}
\section{Experiments}

\subsection{Datasets}
\textbf{CTW1500} is a challenging dataset for curve text detection. It consists of $1,000$ training images and $500$ test images. The text instances are annotated by $14$-polygon.
\\ \textbf{Total-Text} \cite{ch2017total} consists of $1,255$ training images and $300$ test images, including horizontal, curve and multi-oriented texts.
\\ \textbf{MSRA-TD500} \cite{yao2012detecting} is a multi-language dataset. It consists of $3,000$ training images and $200$ test images. The texts instances are annotated with quadrilaterals.
\\ \textbf{SynthText-150K} \cite{liu2020abcnet} is a synthetic dataset for arbitrary-shaped scene text, containing nearly $150k$ images that include curved and multi-oriented texts.

\subsection{Implementation Details}

We set the text scale ranges of P3, P4 and P5 in FPN to $[0, 0.25]$, $[0.2, 0.65]$, $[0.55, 1]$ of the image size, respectively. The dimension $M$ of eigenanchor space is set to $14$. The sampling number $K$ in the sparse assignment scheme is $3$. The loss weight $\lambda$ in Eq.~\eqref{eq-Selection-metric} is $2$. When training from scratch, we adopt stochastic gradient descent (SGD) to optimize our network with the weight decay of $0.0005$ and the momentum of $0.9$. For each dataset the training is conducted on its own training set. We set the training batch size to $4$ and train $500$ epochs, with the learning rate initialized at $0.001$. For more comprehensive comparisons, we pre-train our model on SynthText-150K for $5$ epochs, and then fine-turn $300$ epochs for all datasets. The employed data augmentation includes random rotation, random scale, random flip and random crop.

In the testing stage, we set the short sides of the test image to $704$, $1000$, $800$ for CTW1500, Total-Text and MSRA-TD500, respectively. 
To maintain the original aspect ratio, we resize the long side accordingly. 
All experiments are carried out on an NVIDIA RTX3090 GPU.

\begin{table}[t]
\begin{tabular}{@{\extracolsep{11pt}}ccccc}
\hline
Method   &  R &  P     &  F    &  FPS     \\ \hline
Dense  & $85.6$   & $89.7$ & $87.6$ & $14.6$ \\
Sparse  & $84.8$   & $89.2$ & $86.9$ & $\mathbf{22.3}$ \\ 
Dual  & $\mathbf{85.7}$   & $\mathbf{90.5}$ & $\mathbf{88.1}$ & $22.1$ \\ 
\hline
\end{tabular}
\centering
\caption{Performance of LRANet with different positive sample assignment methods on Total-Text.}
\label{tabel-ablation-dense-sparse-dual}
\end{table}

\begin{table}[]
\centering
\begin{tabular}{@{\extracolsep{12pt}}ccccc}
\hline
Dim   &  R &   P    &  F    &  IoU     \\ \hline
$10$  & $84.6$   & $87.9$ & $86.2$ & $95.1$ \\ 
$14$   & $84.9$   & $\mathbf{89.1}$          & $\mathbf{86.9}$ & $98.0$ \\ 
$18$   & $\mathbf{85.2}$   & $88.6$          & $\mathbf{86.9}$ & $\mathbf{98.8}$  
\\ \hline
\end{tabular}
\caption{Experimental results on CTW1500 for different eigenanchor space dimension.}
\label{tabel-dim}
\end{table}

\subsection{Ablation Study}
To evaluate the effectiveness of the proposed components, we conduct ablation experiments on CTW1500 and Total-Text, with no pre-training used by default.
\\ \textbf{Different Data Representation} The current parameterized text shape methods can be divided into three types: parameterized text contour \cite{zhu2021fourier}, parameterized text mask \cite{su2022textdct}, and parameterized top and bottom sides \cite{liu2020abcnet}. To explore the influence of different data representations for LRA, we construct the matrix $\mathbf{A}$ in Eq.~\eqref{eq-A=u...} based on each of the above three types, and set $M$ to $14$. Qualitative results are shown in Fig.~\ref{fig-mask-line-polygon}. The mask-driven method contains redundant information, making it difficult to extract text shapes exactly. In comparison, the side-driven and contour-driven methods can extract text shape information more accurately, with the contour-driven method excelling at capturing complex text shape information. Consistent with this observation, the results in Table~\ref{tabel-mask-side-contour} indicate that the mask-driven method has poor representation quality, while the contour-driven method achieves the best representation quality and detection performance. 
\\ \textbf{Effectiveness of Dual Assignment Scheme}
We compare the proposed dual assignment scheme with both sparse and dense assignments. To make a fair comparison with dense assignment, we adopt the dense matching strategy from \cite{wang2022tpsnet} as our baseline. As shown in Table~\ref{tabel-ablation-dense-sparse-dual}, our dual assignment achieves a faster inference speed ($22.1$ vs. $14.6$) due to the notable reduction in redundant prediction, while yielding a $0.5\%$ improvement in F-measure compared to dense assignment. Moreover, as shown in Fig.~\ref{fig-convergence}, compared to using only sparse assignment, our dual assignment achieves faster convergence and higher accuracy due to receiving ample supervised signals. It attains a similar FPS to sparse assignment but a better F-measure, see Table~\ref{tabel-ablation-dense-sparse-dual}.
\\ \textbf{Different Number of Sparse Sampling}
We further explore the influence of the sparse sampling number $K$ on model performance. As $K$ increases to $3$, the F-measure is improved from $87.3\%$ to $88.1\%$, and due to the small sparse sampling number, the impact of NMS on inference speed remains limited. However, as $K$ continues to increase, there is no further performance improvement, and the impact of NMS on inference speed is gradually obvious. This observation is consistent with the comparison results against dense assignment in Table~\ref{tabel-ablation-dense-sparse-dual}, indicating that with sufficient supervision, a sparse positive sampling is sufficient.
\begin{table}[]
\centering
\begin{tabular}{@{\extracolsep{10pt}}cccccc}
\hline
$K$   &  R &  P     &  F    &  NMS &FPS    \\ \hline
$5$   & $85.5$   & $90.1$          & $87.8$ & $\checkmark$ & $20.6$ \\  
$3$  & $\mathbf{85.7}$   & $\mathbf{90.5}$ & $\mathbf{88.1}$ &$\checkmark$ & $22.1$ \\ 
$1$  & $85.4$   & $89.2$ & $87.3$ &$\checkmark$ & $22.9$ \\ 
$1$  & $83.8$   & $89.6$ & $86.6$ & $\times$ & $\mathbf{23.5}$ \\ 
\hline
\end{tabular}
\caption{Performance of LRANet with different number of sparse samples on Total-Text.}
\label{table-ablation-number-of-k}
\end{table}
\begin{table}[]
 \centering
\begin{tabular}{@{\extracolsep{0.1pt}}c|c|ccccc} 
\hline
Dataset    & Input    & Ext     & R     & P     & F & FPS    \\ \hline
          & $512$    & $\checkmark$     & $83.9$          & $87.8$          & $85.8$   & $\mathbf{62.8}$       \\
\multirow{1.2}{*}{ CTW1500}   & $608$  & $\checkmark$ & $84.8$          & $89.1$          & $86.9$   & $50.2$       \\
          & $704$  & $\checkmark$ & $\mathbf{85.5}$          & $\mathbf{89.4}$          & $\mathbf{87.4}$   & $37.2$       \\ \hline
           & $608$   & $\checkmark$     & $81.3$      & $\mathbf{90.4}$          & $85.6$  & $\mathbf{48.2}$        \\
\multirow{1.2}{*}{Total-Text}  & $800$ & $\checkmark$ & $86.6$          & $88.9$          & $87.7$    & $32.6$      \\
          & $1000$  & $\checkmark$ & $\mathbf{87.8}$          & $90.3$          & $\mathbf{89.0}$   & $22.1$       \\ \hline
\end{tabular}
 \caption{Performance of LRANet with different input sizes.}
 \label{tabel-shape}
\end{table}
\begin{table}[]
\centering
\begin{tabular}{@{\extracolsep{6pt}}lccccc}
\hline
Method     &  Dim    &  IoU     \\ \hline
Chebyshev \cite{wang2020textray}   & $44$ & $83.6$\\ 
DCT \cite{su2022textdct}           & $32$ & $88.5$ \\ 
Fourier \cite{zhu2021fourier}           & $22$ & $91.5$\\
Bezier \cite{liu2020abcnet}       & $16$ & $97.6$\\
TPS \cite{wang2022tpsnet}           & $22$ & $97.9$\\ 
 \hline
\textbf{LRA}  & $\mathbf{14}$ & $\mathbf{98.0}$ \\ 
\hline
\end{tabular}
\caption{Comparison with different parameterized text shape methods on CTW1500.}
\label{tabel-compare-dim-iou}
\end{table}
\\ \textbf{Dimension of Eigenanchor Space}
We also conduct experiments to verify the influence of the eigenanchor space dimension, i.e., $M$. The results are listed in Table~\ref{tabel-dim}. At $M=10$, the eigenanchor space can roughly represent the original space. As $M$ increases to $14$, the IoU increases from $95.1$ to $98.0$, and the F-measure gains an $0.8\%$ improvement. With the further increase of $M$, there is no significant improvement in the representation quality. The F-measure is almost unchanged. Therefore, we set $M=14$ to balance training complexity and representation quality.
\\ \textbf{Different Input Image Sizes}
To demonstrate the trade-off between speed and accuracy, we evaluate our model with different length of short sides. The results are shown in Table~\ref{tabel-shape}, revealing an ascending F-measure trend as the short side extends, accompanied by a reduction in FPS.  
\begin{table*}[]
\centering
\begin{tabular}{l|c|ccc|ccc|cccc}
\hline
\multirow{2}{*}{Method} & \multirow{2}{*}{Ext} & \multicolumn{3}{c|}{MSRA-TD500} & \multicolumn{3}{c|}{Total-Text} & \multicolumn{4}{c}{CTW1500}  \\       \cline{3-12}      &                      & R      & P      & F        & R          & P       & F      & R        & P        & F  & FPS     \\ \hline 
DB  \cite{liao2020Real}        & $\checkmark$        & $79.2$         & $91.5$      & $84.9$     & $82.5$      & $87.1$       & $84.7$      & $80.2$     & $86.9$        & $83.4$ & $33.5$          \\
TextBPN   \cite{zhang2021adaptive}       & $\checkmark$        & $84.5$                 & $86.6$      & $85.6$   & $85.2$      & $\mathbf{90.8}$       & $87.9$     & $83.6$                 & $86.5$      & $85.0$  & $18.1$         \\
DB++  \cite{liao2022real}        & $\checkmark$        & $83.3$                 & $91.5$      & $87.2$    & $83.2$      & $88.9$       & $86.0$     & $82.8$                 & $87.9$      & $85.3$   & $\mathbf{38.3}$          \\
TextPMs \cite{zhang2022arbitrary}        & $\checkmark$        & $\mathbf{87.0}$                 & $91.0$      & $88.9$      & $87.7$      & $90.0$       & $88.8$      & $83.8$                 & $87.8$      & $85.7$    & $14.4$         \\ 
FSG \cite{tang2022few}        & $\checkmark$        & $84.8$                 & $91.6$      & $88.1$    & $85.7$      & $90.7$       & $88.1$     & $82.4$                 & $88.1$      & $85.2$   & -         \\  
LeafText  \cite{yang2023text}        & $\checkmark$        & $83.8$                 & $92.1$      & $87.8$    & $84.0$      & $\mathbf{90.8}$       & $87.3$        & $83.9$     & $87.1$        & $85.5$  & -          \\ \hline
TextRay \cite{wang2020textray}         &  $\checkmark$     & -               & -  & -   & $77.9$     & $83.5$      & $80.6$       & $80.4$       & $82.8$     & $81.6$      & -             \\
FCENet  \cite{zhu2021fourier}        &  -        & -                 & -  & -   & $82.5$     & $89.3$      & $85.8$     & $83.4$                 & $87.6$  & $85.5$  & -                 \\
PCR  \cite{dai2021progressive}        & $\checkmark$        & $83.5$                 & $90.8$      & $87.0$    & $82.0$      & $88.5$       & $85.2$      & $82.3$                 & $87.2$      & $84.7$  & -         \\
ABCNet V2 \cite{liu2021abcnet}          & $\checkmark$       & $81.3$                 & $89.4$      & $85.2$   & $84.1$      & $89.2$       & $87.0$  &  $83.8$                 & $85.6$      & $84.7$  & -        \\
TextDCT \cite{su2022textdct}         &  $\checkmark$        & -                 & -  & -   & $82.7$     & $87.2$      & $84.9$     & $85.3$            & $85.0$  & $85.1$    & $19.5$          \\ 
TPSNet \cite{wang2022tpsnet}        & -        & -                 & -      & -     & $84.0$      & $89.2$       & $86.6$        & $83.7$             & $88.1$      & $85.9$      & $17.9$         \\ 
TPSNet \cite{wang2022tpsnet}        & $\checkmark$        & -                 & -      & -    & $86.8$      & $89.5$       & $88.1$    & $85.1$                 & $87.7$      & $86.4$    & $17.9$        \\

CT-Net  \cite{shao2023ct}        & -        & $80.4$                 & $89.8$      & $84.8$     & $83.6$      & $89.2$       & $86.3$       & $82.7$     & $87.9$ 
& $85.2$  & $13.6$           \\ 
\hline
\textbf{LRANet}          & -       & $85.3$                 & $89.1$      & $87.2$     & $85.7$      & $90.5$       & $88.1$        & $84.9$     & $89.1$        & $86.9$   & $37.2$ \\ 
\textbf{LRANet}          & $\checkmark$       & $86.3$                 & $\mathbf{92.3}$      & $\mathbf{89.2}$     & $\mathbf{87.8}$      &$90.3$       & $\mathbf{89.0}$      & $\mathbf{85.5}$     & $\mathbf{89.4}$        & $\mathbf{87.4}$   & $37.2$ \\ \hline
\end{tabular}
\caption{Comparison with previous methods on CTW1500, Total-Text and MSRA-TD500. Ext means using external training data to pretrain the model. 
All listed FPS is obtained from the same NVIDIA RTX3090 GPU.}
\label{tabel-compare-previous}
\end{table*}

\begin{figure*}[t]
\centering
\includegraphics[width=\textwidth]{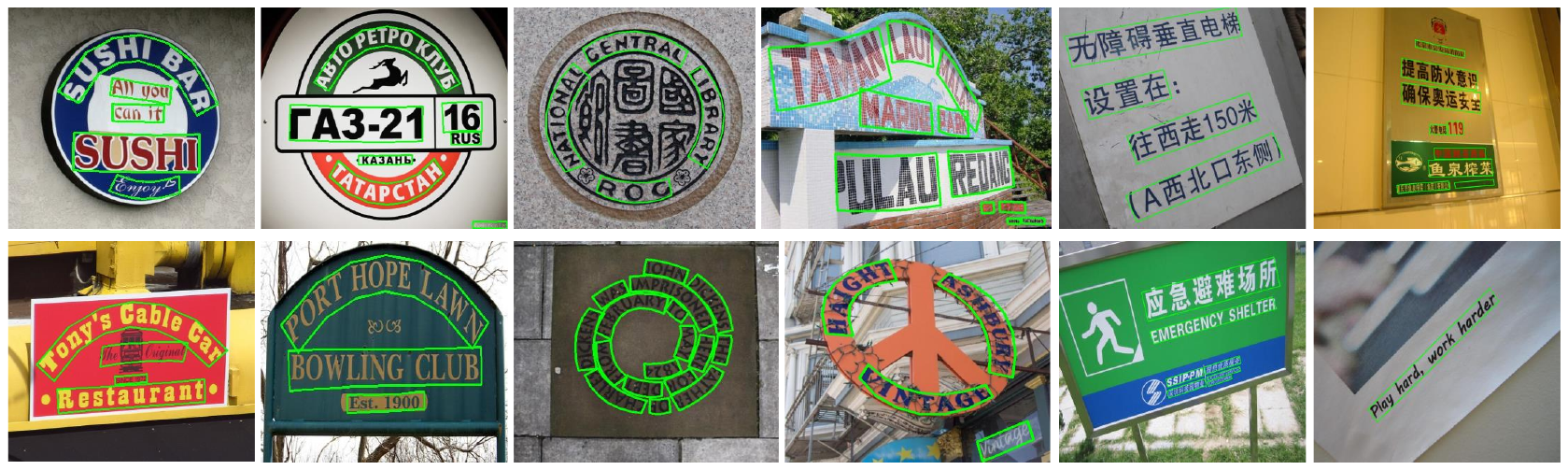}
\caption{Text detection visualizations on the evaluated datasets.}
\label{fig-visual}
\end{figure*}

\subsection{Comparison with State-of-the-art Methods}

We compare LRANet with previous methods on Total-Text, CTW1500, and MSRA-TD500. The results are given in Table~\ref{tabel-compare-previous}. LRANet consistently has top-tier performance across the three datasets. Compared to the segmentation-based methods, LRANet achieves competitive results even without pre-training, demonstrating its robustness to noise resistance. Compared to other parameterized text shape methods, LRA achieves superior results with fewer parameters, as shown in Table~\ref{tabel-compare-dim-iou}. The results reveal the significance of learning text-specific shape information, and our LRA effectively bridges this gap compared to existing solutions. It is observed that, when compared to the regression-based methods, LRANet shows advantages in terms of inference speed, while also maintaining leading performance. On one hand, the speed advantages come from the dual assignment. It provides sufficient supervised signals for model training, endowing the model with the capability of selecting high-quality positive samples at a single glance. On the other hand, the performance advantages come from LRA. For example, although TPSNet introduces a border alignment loss to alleviate the instability in its predicted TPS points, LRANet outperforms it by $1.0\%$ and $1.5\%$ in terms of F-measure on CTW1500 and Total-Text, respectively, even without intricate loss design. This is mainly attributed to the more advanced shape-relevant representation of LRA. Some detection visualizations are shown in Fig.~\ref{fig-visual}. LRANet performs well on long, curve, and perspective texts.

\section{Conclusion}
In this paper, we present LRANet, an accurate and efficient arbitrary-shaped text detector. LRANet is featured by proposing an eigenvector-based reconstruction to effectively leverage the text-specific shape information. It accurately regresses the text contour with fewer parameters. Meanwhile, a dual assignment scheme that well integrates both dense and sparse assignments is developed for efficient inference. Experiments conducted on public benchmarks basically verify the proposed LRANet, where top-ranked performance and extremely fast inference speed are simultaneously observed. Given its effectiveness and efficiency, we are also interested in extending LRANet to text spotting in future.

\section{Acknowledgements}
The work was supported by National Key R\&D Program of China (No. 2020AAA0104500) and the National Natural Science Foundation of China (No. 62172103).

\appendix
\section{More Experiments of LRA}
\subsection{Generalization Evaluation of LRA}

To verify the generalization of LRA, we extract the first $14$ eigenanchors from the accessible synthetic dataset SynthText-150K and evaluate their performance on real datasets. As shown in  Table~\ref{tabel-ablation-generality-lra}, 
compared to eigencontours extracted from their respective training datasets, eigencontours obtained from SynthText-150K consistently maintain good representation quality and model performance in real-world scene text, indicating the generalization of our LRA.

\begin{table}[h]
 \setlength\tabcolsep{2pt}
 \renewcommand\arraystretch{1.2}
\centering
\caption{Generalization Evaluation of LRA.}
\label{tabel-ablation-generality-lra}
\begin{tabular}{c|cccccc} 
\hline
   Eigenanchor Space  & Dataset        & R     & P     & F & IoU    \\ \hline
\multirow{2}{*}{SynText-150K}   & CTW1500 & $85.3$          & $88.7$          & $86.9$   & $97.8^{(\downarrow0.2)}$       \\
          & Total-Text  & $85.2$          & $91.1$          & $88.1$   & $98.3^{(\downarrow0.1)}$       \\ \hline
\end{tabular}
\end{table}

\subsection{Robustness Evaluation of LRA}

\begin{figure}[!h]
    \centering   
    \includegraphics[width=0.47\textwidth]{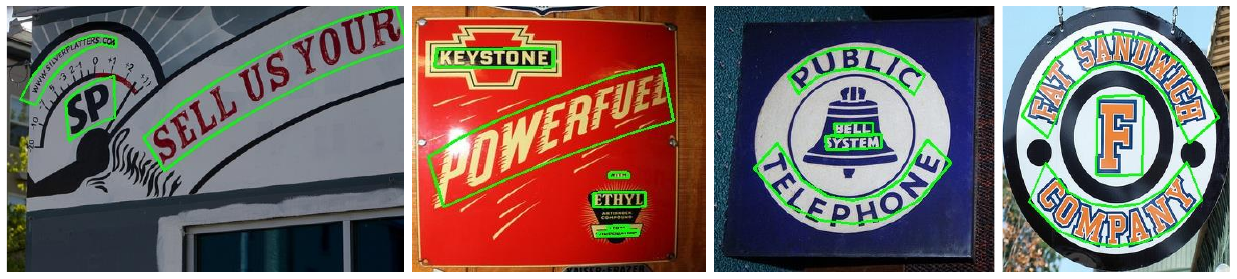}
    
    \caption{The results of only using the top six eigenanchors ($u_1, \dots, u_6$) for LRA decoding.}
\label{fig-six_engen}
\end{figure}

Table~\ref{table-dimension-reduction} shows the results of our LRANet using different number of eigenanchors. By setting the LRA coefficients for the last $4$ dimensions to zero and employing only the coefficients from the top $10$ dimensions along with their corresponding eigenanchors for LRA decoding, the F-measure decreases by a mere $0.2$. Continuing this trend by setting the coefficients of the last $8$ dimensions to zero, the F-measure again drops slightly. Moreover, when we further exclude $u_1$, which capture the most important information in the original space, and only use $u_2, \dots, u_6$ for LRA decoding. Surprisingly, even with just five parameters for contour representation, the performance still surpasses TextRay (Wang et al. 2020) with $44$ parameters for contour representation ($81.1$ vs. $80.6$ in F-measure). This indicates that our LRA has strong representation capacity and robustness. 
Furthermore, as shown in Fig.~\ref{fig-six_engen}, multi-directional and extremely curved text can be accurately fitted using only the first six dimensions of the LRA coefficients.

\begin{table}[!h]
\setlength\tabcolsep{4pt}
\centering
\caption{Robustness analysis of LRA on Total-Text.  The baseline model is our LRANet with $M=14$. Disturb refers to adjusting the predicted LRA coefficient of a certain dimension to a disturbance value.}
\label{table-dimension-reduction}
\begin{tabular}{llcc}
\hline
Method & Dim & Disturb   & F-measure \\
\hline
Basiline+ & Last-$4$  & 0  & $87.9^{(\downarrow0.2)}$
\\
Basiline+ & Last-$8$  & 0 & $87.2^{(\downarrow0.9)}$
\\
Basiline+ & Last-$8$ & 1/2 & $87.9^{(\downarrow0.2)}$
\\
Basiline+ & Last-$8$ & 2  & $87.6^{(\downarrow0.5)}$
\\
Basiline+ &Last-$8$ \& Top-$1$ & 0 & $81.1^{(\downarrow7.0)}$
\\

\hline
\end{tabular}
\end{table}

\subsection{Different Number of Contour Point}

Generally, as the number of contour points increases, the representation quality of parameterized text shape methods  tends to decrease. To investigate the impact of contour points on LRA, we conduct ablation experiments on CTW1500. As shown in Table~\ref{tabel-ablation-N}, the representation quality of LRA remains consistent despite an increase in the number of contour points. This indicates that our LRA is well-suited for scenes with more refined text contour annotations. In addition, more contour points may raise training difficulty, resulting in a slight performance decrease.

\begin{table}[!h]
\setlength\tabcolsep{12pt}
\centering
\caption{Experimental results on CTW1500 for different text contour points.}
\label{tabel-ablation-N}
\begin{tabular}{ccccc}
\hline
$N$  &  R &   P    &  F    &  IoU     \\ \hline
$14$  & $84.9$   & $\mathbf{89.1}$ & $\mathbf{86.9}$ & $\mathbf{98.0}$ \\ 
$28$   & $\mathbf{85.4}$   & $88.4$          & $\mathbf{86.9}$ & $\mathbf{98.0}$ \\ 
$56$   & $84.1$   & $88.8$   & $86.4$ & $\mathbf{98.0}$  
\\ \hline
\end{tabular}
\end{table}

\section{More Visualizations}
\subsection{Visualization of Eigenanchors}

\begin{figure*}[t]
\centering
\includegraphics[width=\textwidth]{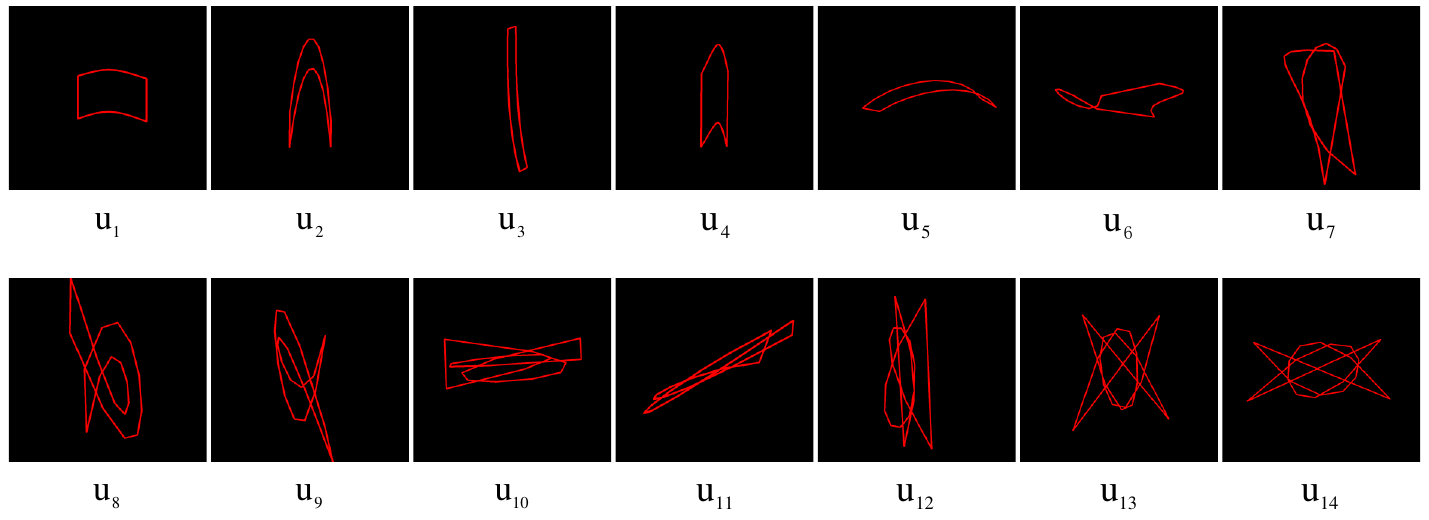}
\caption{Visualization of the $14$ eigenanchors for the SynthText-150K dataset.}
\label{fig-visual-lra}
\end{figure*}

As shown in Fig.~\ref{fig-visual-lra}, 
the first three eigenanchors mainly capture the overall information of the text shape, with each one focusing on the rough outline, curvature state, and orientation, respectively. As the singular value gradually decreases, the subsequent eigenanchors are more focused on capturing detailed information of the text shape,  resulting in their shapes becoming more complex.

\subsection{Qualitative Comparisons}
\begin{figure*}[!t]
\centering
\includegraphics[width=\textwidth]{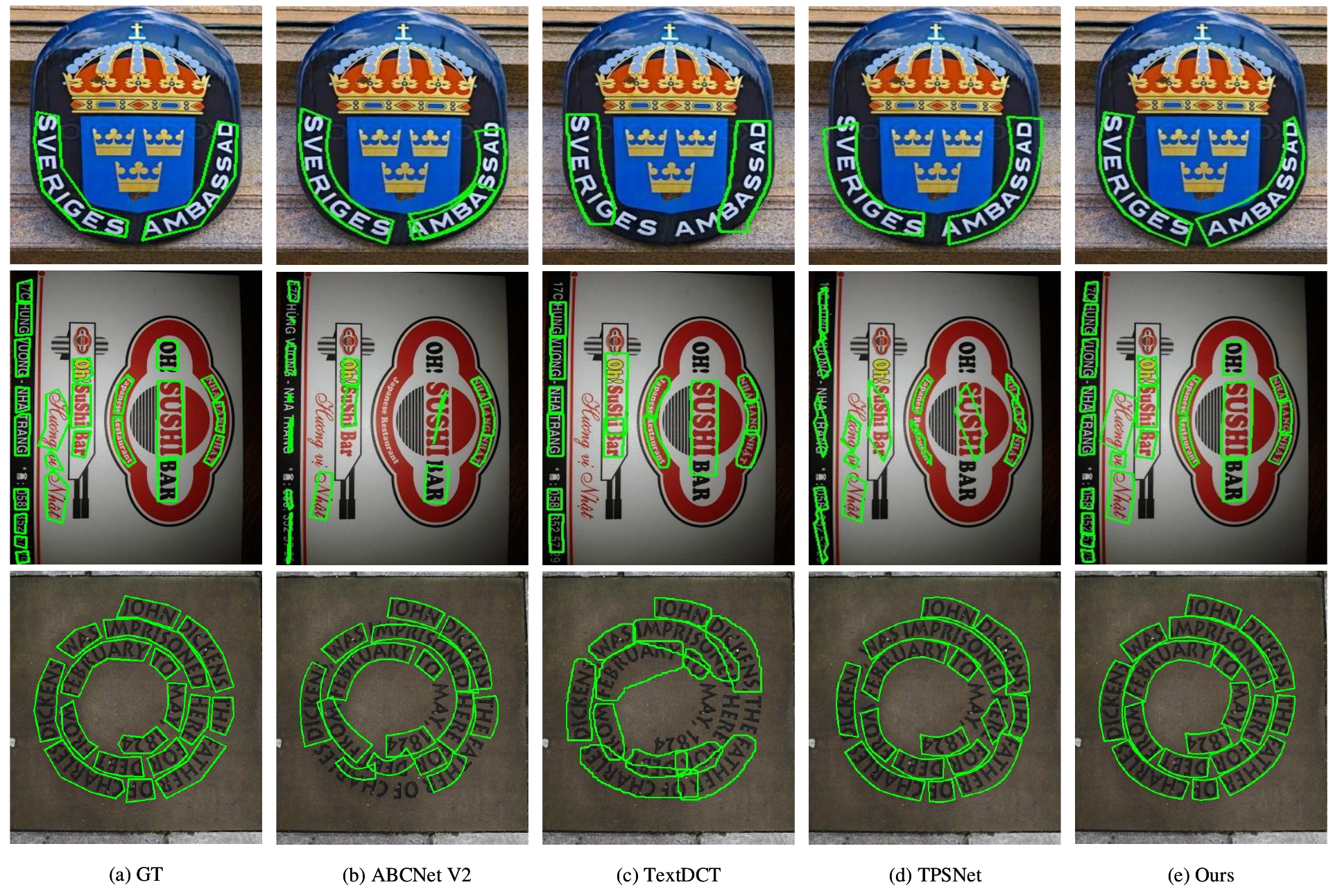}
\caption{Qualitative comparisons of ABCNet V2 (Liu et al. 2021), TextDCT (Su et al. 2022), TPSNet (Wang et al. 2022) and our LRANet in Total-Text.}
\label{fig-compare-beizer-tps-dct}
\end{figure*}

To visually demonstrate the effectiveness of LRA, we qualitatively compared it with the parameterized top and bottom sides method ABCNet V2, the parameterized text masks method TextDCT and the parameterized text contours method TPSNet. Notably, all these methods adopt a single-stage anchor-free framework, including our LRANet. As illustrated in Fig.~\ref{fig-compare-beizer-tps-dct}, our LRANet performs well on vertical, curve, and dense texts, even significant challenges are presented.

\bibliography{aaai24}
\end{document}